\makeatletter\def\graphicscache@inhibit{true}\makeatother

\documentclass[letterpaper, 10 pt, conference]{ieeeconf}  %

\IEEEoverridecommandlockouts                              %

\usepackage{graphics} %
\usepackage{epsfig} %
\usepackage{amsmath} %
\usepackage{subfig}
\usepackage{hyperref}
\usepackage{wrapfig}
\usepackage{pgfplots}
\usepackage[capitalize]{cleveref}
\usepackage{tabularx} \usepackage{booktabs} \usepackage{multirow}

\usepackage{tikz}
\usetikzlibrary{arrows.meta,
                arrows,
                positioning,
                shadows}
\usetikzlibrary{arrows}
\usetikzlibrary{positioning,calc}
\usetikzlibrary{decorations.pathreplacing}
\usetikzlibrary{decorations.markings}
\usetikzlibrary{decorations.pathmorphing}
\usetikzlibrary{calligraphy}
\usetikzlibrary{fit}
\usetikzlibrary{shapes.arrows}
\usetikzlibrary{shapes.geometric}
\usetikzlibrary{matrix}
\usetikzlibrary{spy}
\usepackage{pgfplots}
\usepgfplotslibrary{groupplots}
\usetikzlibrary{fit,shapes.callouts,shapes.geometric,backgrounds,positioning,arrows.meta}
\graphicspath{{./figures/}}

\makeatletter
\tikzdeclarecoordinatesystem{rel}{%
	\tikzset{cs/.cd,x=0pt,y=0pt,#1}%
	\pgfpointlineattime{(\tikz@cs@x / 100)}%
	{\pgfpointanchor{\tikz@pp@name{\tikz@cs@node}}{south west}}%
	{\pgfpointanchor{\tikz@pp@name{\tikz@cs@node}}{south east}}%
	\edef\tikz@cs@x{\the\pgf@x}%
	\pgfpointlineattime{(\tikz@cs@y / 100)}%
	{\pgfpointanchor{\tikz@pp@name{\tikz@cs@node}}{south west}}%
	{\pgfpointanchor{\tikz@pp@name{\tikz@cs@node}}{north west}}%
	\pgfpoint{\tikz@cs@x}{\pgf@y}%
}
\makeatother

\usepackage{tensor}
\newcommand{\T}[2]{\ensuremath{\tensor[^{#2\,}]{T}{_{\,#1}}}}
\usepackage[style=ieee,hyperref,natbib=true,backend=bibtex,firstinits,doi=false,%
     mincitenames=1,maxcitenames=2,maxbibnames=99,sorting=none,terseinits=false,hyperref=true]{biblatex}
\bibliography{references}
\renewbibmacro*{bbx:savehash}{}%
\defbibheading{bibliography}[\bibname]{\section*{References}}

\title{\LARGE \bf
Iterative Motion Compensation for Canonical 3D Reconstruction from UAV Plant Images Captured in Windy Conditions
}

\author{
	\begin{tabular}{cccc}
		Andre Rochow & Jonas Marcic & Svetlana Seliunina & Sven Behnke \\[-2pt]
		{\tt\footnotesize rochow@ais.uni-bonn.de} & {\tt\footnotesize s6jsmarc@uni-bonn.de} & {\tt\footnotesize s45sseli@uni-bonn.de} & {\tt\footnotesize behnke@cs.uni-bonn.de}
	\end{tabular}
	\\[6pt]
	\begin{tabular}{c}
		{\small Autonomous Intelligent Systems - Computer Science Institute VI and Center for Robotics, University of Bonn, Germany} \\[0pt]
		{\small Lamarr Institute for Machine Learning and Artificial Intelligence, Germany}
	\end{tabular}
}

\usepackage{tabulary}

\begin{document}

\maketitle
\thispagestyle{empty}
\pagestyle{empty}

\begin{abstract}

3D phenotyping of plants plays a crucial role for understanding plant growth, yield prediction, and disease control. We present a pipeline capable of generating high-quality 3D reconstructions of individual agricultural plants.
To acquire data, a small commercially available UAV captures images of a selected plant. Apart from placing ArUco markers, the entire image acquisition process is fully autonomous, controlled by a self-developed Android application running on the drone’s controller. The reconstruction task is particularly challenging due to environmental wind and downwash of the UAV.
Our proposed pipeline supports the integration of arbitrary state-of-the-art 3D reconstruction methods. 
To mitigate errors caused by leaf motion during image capture, we use an iterative method that gradually adjusts the input images through deformation. Motion is estimated using optical flow between the original input images and intermediate 3D reconstructions rendered from the corresponding viewpoints. This alignment gradually reduces scene motion, resulting in a canonical representation.
After a few iterations, our pipeline improves the reconstruction of state-of-the-art methods and enables the extraction of high-resolution 3D meshes. We will publicly release the source code of our reconstruction pipeline. Additionally, we provide a dataset consisting of multiple plants from various crops, captured across different points in time.

\end{abstract}

\section{Introduction}

In agriculture, phenotyping individual plants is essential for detecting pests, diseases, and assessing growth. 3D reconstruction provides plant scientists with a powerful tool to study plants in greater detail. Unmanned ground vehicles (UGVs) have been used to capture multiple images of a plant simultaneously~\cite{esser2023field}. However, such systems are typically expensive, and need driving access to the plants of interest.
In this work, we focus on data acquisition using commercially available and cost-effective drones, such as the DJI Mini Pro 3.
Accurately reconstructing plants in 3D using UAV imagery is particularly challenging due to downwash generated by the copter, which causes substantial leaf motion.
Compared to UGV solutions~\cite{esser2023field}, images can no longer be captured simultaneously from multiple perspectives.
Several methods based on Neural Radiance Fields (NeRF)~\cite{mildenhall2021nerf} have been proposed to handle dynamic scenes, including Non-Rigid NeRF~\cite{tretschk2021non} and Nerfies~\citep{park2021nerfies}. More recently, 3D Gaussian Splatting~\citep{kerbl20233d} has emerged as a new state-of-the-art approach, effectively replacing NeRF in many 3D reconstruction tasks. Following its success, prior concepts from deformable NeRFs have been adapted to 3D Gaussian Splatting to handle dynamic scenes~\citep{pumarola2021d, yang2024deformable}. These methods typically model the complete motion within a scene, enabling interpolation not only across viewpoints but also over time.
In contrast, our goal is canonical 3D reconstruction for plant phenotyping. Therefore, we model motion solely for the purpose of compensating for it.

We present an approach that can be combined with any 3D reconstruction method and iteratively aligns the input images into a canonical configuration using optical flow (see \cref{fig:pipe:3d}). Motion is compensated by selectively deforming the original input images. In the first iteration, the raw images are used directly. In subsequent iterations, deformations of the original images are computed by rendering the intermediate 3D reconstruction from the input viewpoints and estimating the optical flow between these renderings and the original inputs. This flow is then used to warp the input images, progressively reducing scene motion.
We demonstrate that our method significantly improves the performance of various 3D reconstruction algorithms in the presence of motion.

\begin{figure}[t]
	\centering
	\includegraphics[width=\linewidth]{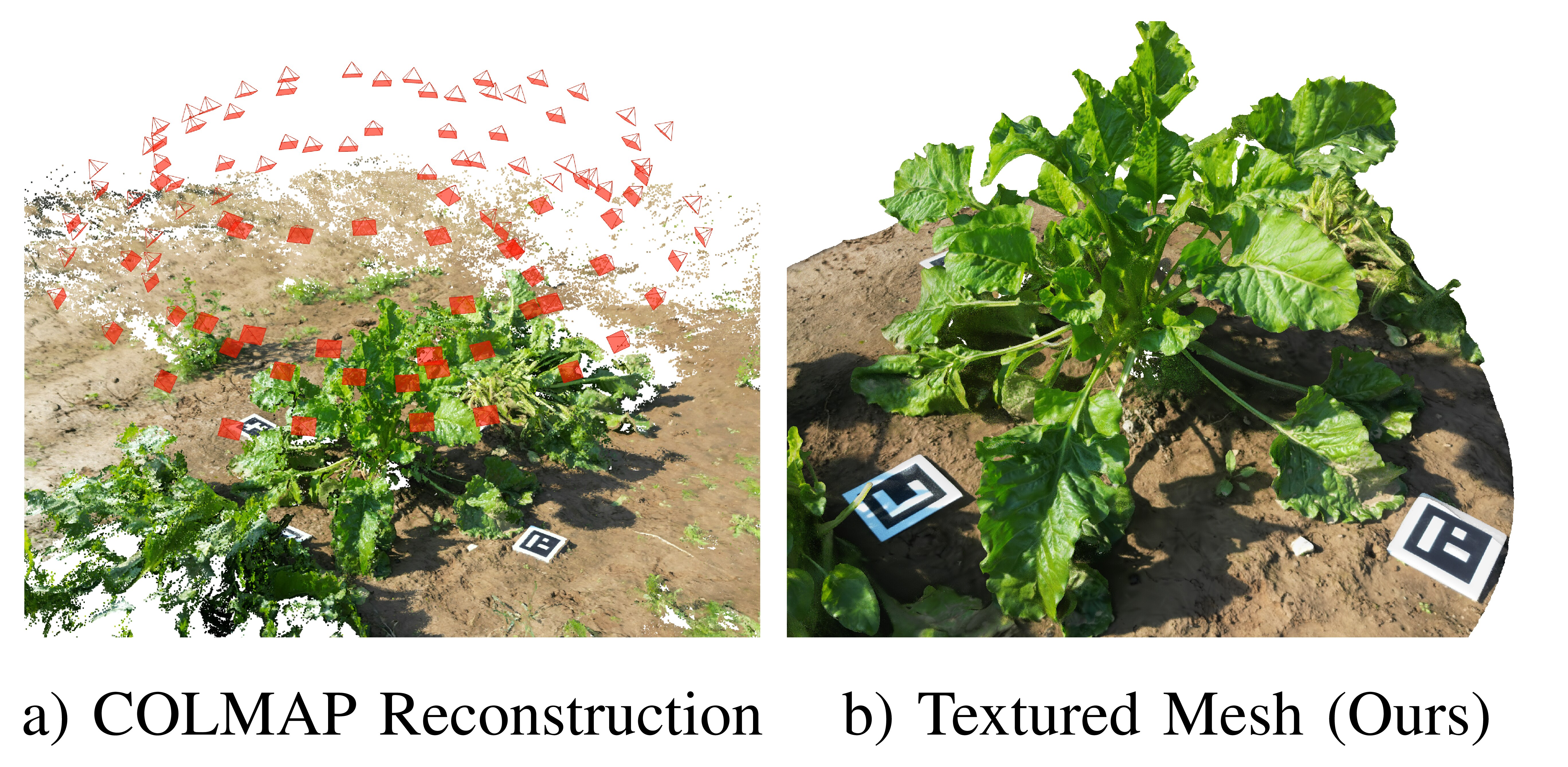}
	\caption{a): Aligned (dense) scene reconstruction using COLMAP~\cite{schoenberger2016sfm}, including the estimated camera poses. b): Textured mesh extracted after 100 iterations with our proposed method (3D Gaussian Splatting~\cite{kerbl20233d} was used as the baseline).} \label{fig:teaser}
\end{figure}

We automate the image capture process to eliminate the tedious and time-consuming task of manual data collection, thus reducing the need for human involvement. The primary challenge here is localization from the limited amount of sensor data provided by the UAV. We therefore used fiducial markers placed around the plant to be captured. The UAV's state is estimated using an extended Kalman Filter. The objective was to reach an adjustable set of waypoints and collect the visual data. Collected visual data combined with the information about marker type can later be used to enforce the correct scale of the reconstructed plant using COLMAP~\cite{schoenberger2016sfm}.

Our contributions can be summarized as follows: 

\begin{enumerate}
    \item[(i)] Autonomous image capturing pipeline for commercially available UAVs.
    \item[(ii)] Dataset containing images of different types of plants at different stages of growth, including the camera parameters.
	\item[(iii)]  A 3D reconstruction pipeline that builds on arbitrary 3D reconstruction baseline methods and iteratively removes motion from the scene via optical flow compensation.
	\item[(iv)] A detailed evaluation that highlights that our method greatly improves the  results of the baseline methods.
\end{enumerate}

\section{Related Work}

\paragraph{UAV Localization and Flight Control} %
UAV localization and path following are necessary parts of many applications and were, therefore, extensively researched over the years. The approaches used differ significantly for indoor and outdoor scenarios. Most of the outdoor applications rely on global navigation satellite systems (GNSS), such as the global positioning system (GPS), combined with an inertial navigation system (INS) inside a sensor fusion framework for pose estimation~\cite{nemra2010robust},~\cite{zhang2018intelligent}. Other authors also combine GNSS data with other relative positioning systems~\cite{mao2007design}. The idea is that the relative position or INS provides short-term accurate data, but will drift in the long term. On the other hand, GNSS does not suffer from error accumulation, but has a big error margin and provides data at a lower rate.

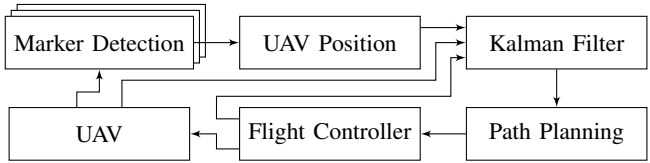
\begin{figure}[tb!]
	\centering
	\begin{tikzpicture}[auto,
	node distance = 5mm and 6mm,
	N/.style = {draw, fill=white, minimum height=7mm, minimum width=24mm, align=right, font=\small},
	dcs/.style = {double copy shadow, shadow xshift=2pt, shadow yshift=-2pt, font=\small},
	>=latex'
	]
	\node[N, dcs] (markers)  {\hyperref[sec:marker]{Marker Detection}};
	\node[N, right=of markers]  (camera) {\hyperref[sec:position]{UAV Position}};
	\node[N, right=of camera]  (kalman) {\hyperref[sec:filter]{Kalman Filter}};
	\node[N, below=of markers]  (uav)    {\hyperref[sec:uav]{UAV}};
	\node[N, below=of kalman]  (path) {\hyperref[sec:planning]{Path Planning}};
	\node[N, left=of path]  (controller) {\hyperref[sec:controller]{Flight Controller}};
	\draw [->] (markers) -- (camera);
	\draw [->] ([xshift= -3 mm]uav.north) |- ([yshift= 2 mm]uav.north) -- (markers.south);
	\draw [->] ([yshift= 2 mm]camera.east) -- ([yshift= 2 mm]kalman.west);
	\draw [->] (kalman) -- (path);
	\draw [->] (path) -- (controller);
	\draw [->] ([yshift= -2 mm]controller.west) -| ([xshift= -3.0 mm]controller.west) -- (uav.east);
	\draw [->] ([yshift= 2 mm]controller.west) -| ([yshift= 5 mm, xshift= -3.0 mm]controller.west) -| ([yshift= -2 mm, xshift= -2 mm]kalman.west) -- ([yshift= -2 mm]kalman.west);
	\draw [->] ([xshift= 3 mm]uav.north) |- ([yshift= -5mm, xshift= -4 mm]kalman.west) |- (kalman.west);
	\end{tikzpicture} 
	\caption{Pipeline of the autonomous capturing method. Firstly, visual data together with UAV orientation, altitude, velocity, and gimbal orientation are received from the UAV. After that, visible markers are extracted, and the UAV position is approximated from them. Data from the UAV, together with the approximated position from markers and the previous command, is sent to the Kalman filter. The current position is approximated and passed to the path planner, which creates a trajectory through desired waypoints. Finally, the motion controller receives the trajectory and sends the velocity command back to the UAV.}
	\label{fig:pipeline}
\end{figure}

Often radio communications reception issues and interferences make GNSS unreliable. One possible modality to mitigate this issue is vision. For outdoor localization, absolute visual localization, which involves matching UAV visual data with reference data, is commonly used~\cite{couturier2021review}. In the absence of reference data, one has to rely on relative visual localization using visual-inertial odometry (VIO) and simultaneous localization and mapping (SLAM)~\cite{gyagenda2022review}.

Visual localization is less effective in environments with low visual feature density or repetitive features. One way of addressing this issues is to use fiducial markers~\cite{kalaitzakis2021fiducial}. They are commonly used for indoor localization~\cite{lim2009real},~\cite{zhenglong2018pose},~\cite{bacik2017autonomous} and for identifying specific places in an outdoor environment, for example, landing zones~\cite{nguyen2017remote}.

\paragraph{3D Reconstruction}
Several 3D reconstruction methods that learn a neural radiance field~\citep{mildenhall2021nerf} for reconstructing a scene from image inputs~\citep{muller2022instant,rosu2023permutosdf,zhang2020nerf,barron2021mip}. \citet{muller2022instant} propose to use a hash-based embedding to improve optimization accuracy and speed. Based on this~\citet{rosu2023permutosdf} replace the voxel-based hash encoding with a permutohedral lattice that allows for faster optimization in higher dimensions. Instead of sampling densities along a ray, a signed distance function (SDF) is optimized, which significantly improves the quality of the mesh extracted from the volume.%

More recently, 3D Gaussian Splatting approaches have been proposed that do not require neural networks to represent scenes. Instead, the scene is modeled as a set of Gaussian primitives, each described by position, orientation, opacity, and shape~\cite{kerbl20233d}. Extensions of this approach include methods that model surfaces instead of volumes, allowing for more accurate mesh extraction~\cite{huang20242d, yu2024gaussian}. Instead of volumetric 3D Gaussians~\citet{huang20242d} use flat 2D Gaussians and~\citet{yu2024gaussian} make use of additionally learned opacity fields. All of these methods assume static scenes. Violations of this assumption lead to blurry reconstructions.%

To model motion in non-rigid scenes, a number of NeRF-based methods~\citep{mildenhall2021nerf} have been proposed~\citep{park2021nerfies, pumarola2021d,tretschk2021non}. These methods introduce an additional neural network to estimate deformation of a canonical volume over time~\cite{park2021nerfies, pumarola2021d, tretschk2021non}.

Similarly, 3D Gaussian Splatting~\citep{kerbl20233d} methods for dynamic scenes incorporate deformation networks to deform Gaussian primitives over time by applying some offset, rotation, and scaling to each point~\cite{wu20244d, yang2024deformable}.%

While these methods support interpolation in time, they are not explicitly optimized to produce a sharp canonical representation of the scene.
In contrast, our approach does not aim to model deformations but reconstruct a static volume from images containing motion. We propose to iteratively compensate motion by deforming input images into a motion-free representation before reconstruction.

\section{Autonomous Capturing Method}
\cref{fig:pipeline} gives an overview of our autonomous image capture. We will discuss each component in the following.

\subsection{UAV} \label{sec:uav}
For our task, we used the DJI Mini 3 Pro UAV. All localization and navigation code was executed on the DJI RC Pro remote controller. We developed a custom Android application that established the communication between the UAV and the controller via the DJI Mobile SDK. Visual data, UAV orientation, altitude, velocity, and gimbal orientation were received and utilized for localization. In return, our controller sent velocity commands to the UAV. GPS data without corrections from another device decreased the accuracy in position estimation, so we chose not to use it. 

\subsection{Marker Detection} \label{sec:marker}
To mark the plant of interest, we use four binary square fiducial markers and position them so that the plant is located in the center. We selected 4$\times$4 ArUco markers and utilized the OpenCV library to detect them and estimate the camera pose relative to the center.

However, since all marker corners are coplanar, the Perspective-n-Point (PnP) pose computation problem becomes ill-posed. A marker can be projected onto the same image pixels from two different camera locations, which creates ambiguity in orientation (\cite{munoz2019spm},~\cite{oberkampf1996iterative}). To resolve this issue, we compare two solutions generated by the PnP algorithm and select the one with the smallest angular difference from the previous pose.

\begin{figure}[h]
    \centering
    \subfloat[\centering maker ring]{{\includegraphics[width=.45\linewidth]{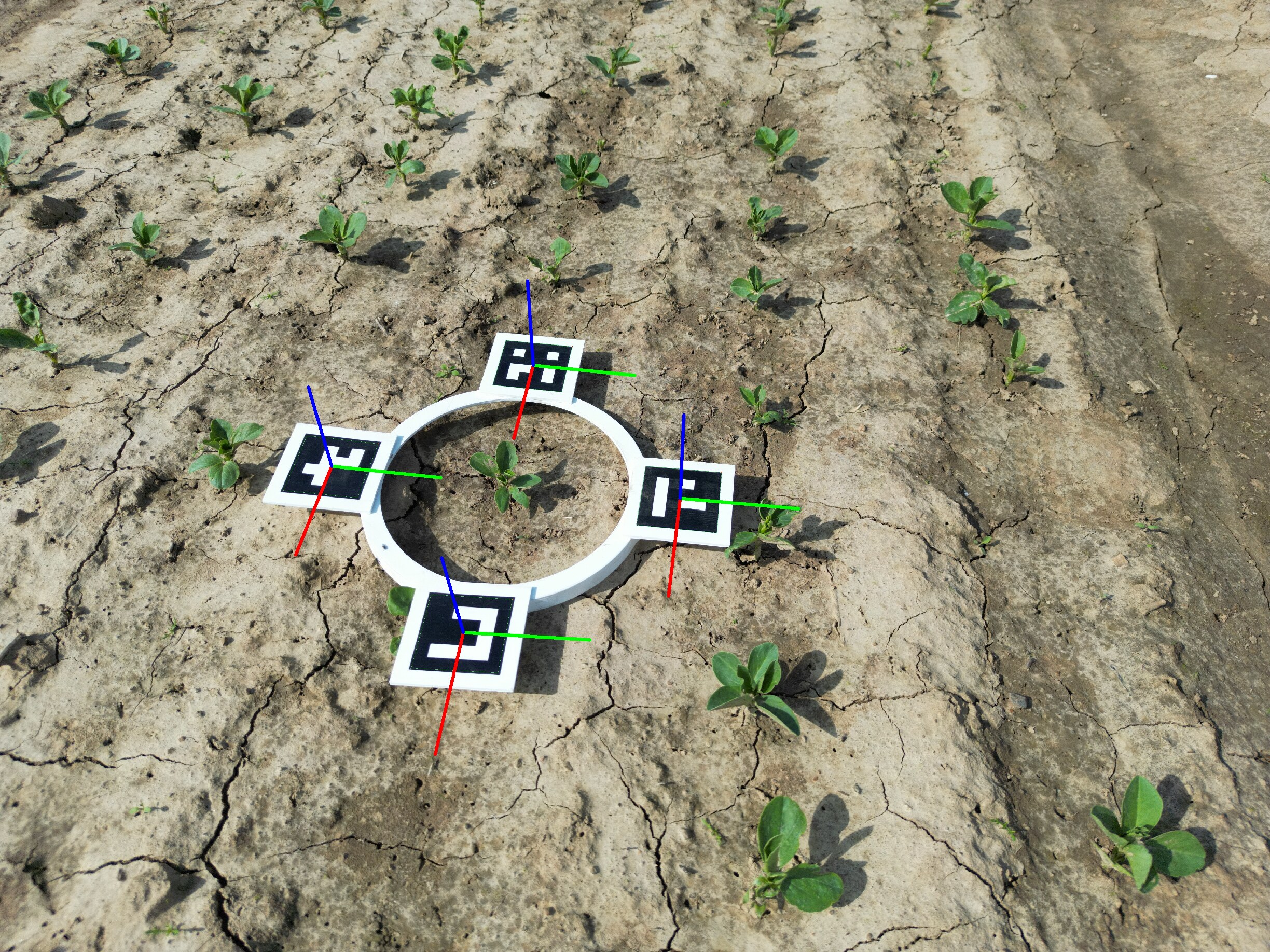} }}%
    \subfloat[\centering separate markers]{{\includegraphics[width=.45\linewidth]{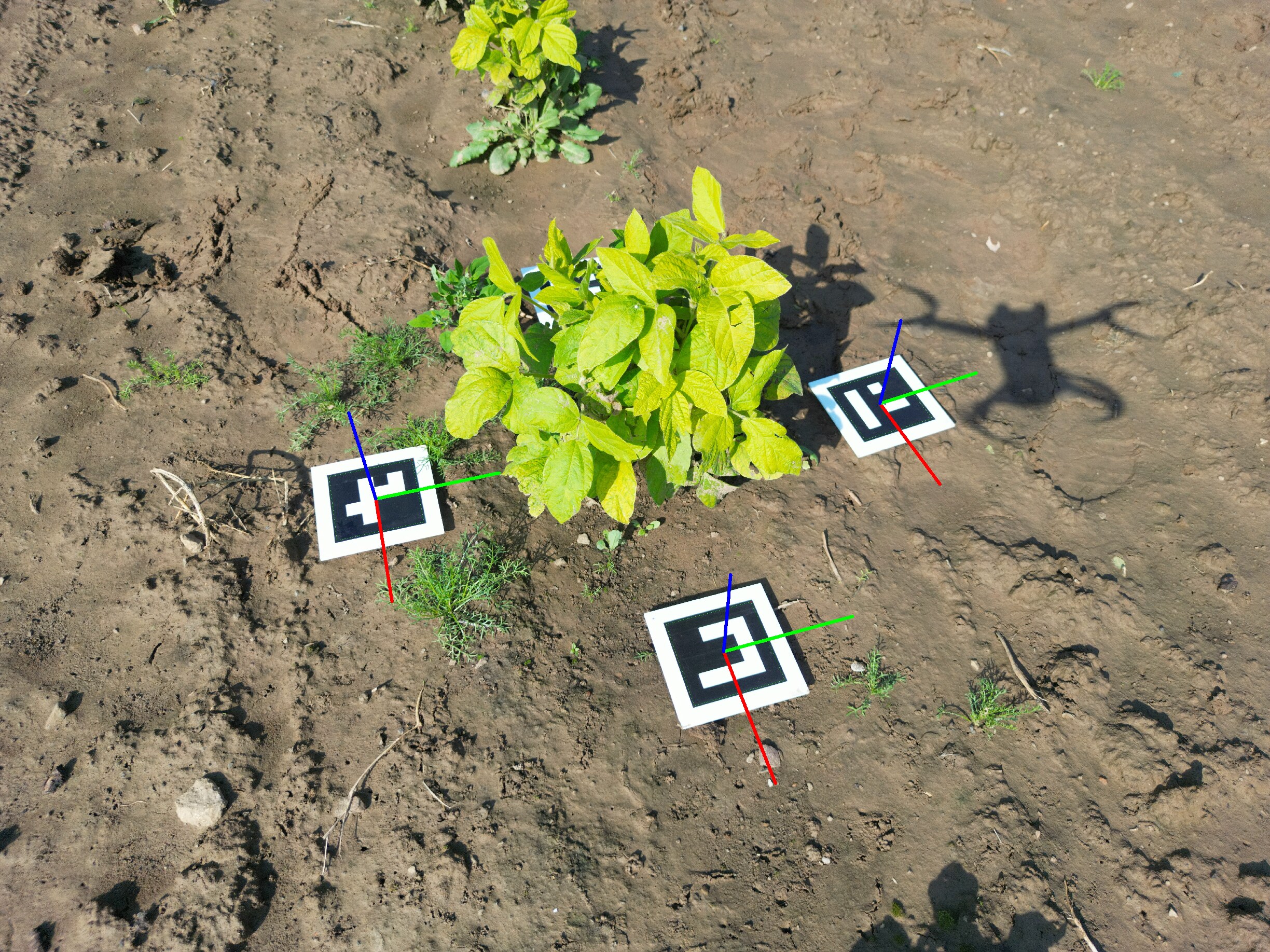} }}%
    \caption{Marker sets with examples of detected markers.}%
    \label{fig:markers}%
\end{figure}

We produced two sets of markers: the first set has smaller markers attached to a ring, while the second set consists of larger, separate markers for bigger plants, as shown in \cref{fig:markers}.

\subsection{UAV Position} \label{sec:position}
The mean position of the markers represents the world coordinate system around which the UAV will navigate. For the ring, we know how the markers are located relative to the center (\T{0..3}{w}). For separate markers, we measure the distance from the first marker to the center and fix \T{0}{w}. The transformations between other markers and the center are estimated on the fly:
\begin{align}
	\T{1..3}{w}= \T{0}{w}\, (\T{0}{c})^{-1}\, \T{1..3}{c},
\end{align} 
where \T{0..3}{c} are current marker positions relative to the camera. To estimate the UAV's pose, we first position the markers so that the plant is centered. After obtaining the marker poses relative to the camera \T{0..3}{c}, we can use the transformation between a marker and the center \T{0..3}{w} to estimate the camera pose relative to the center \T{c}{w}. Finally, by adjusting the orientation by the gimbal tilt, we determine the UAV pose \T{r}{w}.

\subsection{Kalman Filter} \label{sec:filter}
Visual localization using ArUco markers is effective in controlled environments, but its reliability decreases in real-world applications. Changes in light and wind, marker occlusion, or unstable connections can all result in an inability to estimate the correct camera pose. To achieve consistent camera position estimation, we utilize an extended Kalman filter. In addition to pose from visual localization, we combine altitude, attitude, and velocity data during the prediction step. The update step utilizes the velocity command sent to the UAV.

\subsection{Path Planning} \label{sec:planning}
To capture the plant from various angles, we fly multiple circles at different heights around the plant. As we increase the height, the gimbal tilt increases, and the radius of the circles decreases accordingly. By adjusting the number of circles and waypoints, height, and tilt, we can capture high-resolution visual data from all desired angles.

\subsection{Flight Controller} \label{sec:controller}
The goal of motion planning in photo mode is to reach the specified waypoint, defined by its position and orientation. At each step, we calculate the velocity using a feed-forward plus a proportional-integral feedback controller with integral windup and imposed maximal velocity limitation. 
\begin{align}
    \dot{\theta}(t) =  K_f \dot{\theta}_d(t) + K_p \theta_e(t) + K_i \int^t_0 \theta_e(t)dt 
\end{align}
with  $\theta_e(t) = \theta_d(t)-\theta(t)$, where $K_f, K_p, K_i$ are forward, proportional and integral coefficients respectively, $\theta(t)$ - actual trajectory, $\theta_d(t)$ - desired trajectory.

Setpoint ramping is utilized to adjust the waypoint position gradually, preventing abrupt changes when a new waypoint is set. 
The camera's orientation at each step is adjusted to focus on the center.

The performance of the controller depends heavily on the weather conditions. 
Depending on the wind, autonomously capturing a single circle with 25 waypoints took five to ten minutes, resulting in up to 40\,min for the entire plants (four circles).

\subsection{Scene Alignment} %
Many 3D reconstruction pipelines use  Structure-from-Motion (SfM), for example, COLMAP~\cite{schoenberger2016sfm}. An additional advantage of fiducial markers is the ability to deduce the correct scale of the scene from them, which is usually not possible in SfM.

To do so, we extract camera positions from images with markers. If the distance between the center and the marker is unknown, it is set to an arbitrary number. The distance is then modified after all camera poses are extracted so that the center is their mean projected to the ground plane. To enforce the scale, we create the sparse model from visual data and align the proposed camera poses of the model with the actual ones from detected markers. The reconstruction with camera poses is shown in \cref{fig:teaser} (a)).

\subsection{Dataset}
The dataset that we publish consists of high-resolution images of several crop types collected at different stages of their growth. They are grouped by scenes that consist of many images collected at different attitudes, distances to the plant, and camera tilt angles. Most of the time, we flew four circles around the plant with a height increase between the second and third ones and tilt angles of 40, 50, 50, and 60 degrees, respectively. Dataset content is further described in \cref{tab:dataset}. Most scenes consist of 100 images with 25 images per circle, but sometimes the number of pictures per scene varies due to an incomplete flight or the plant not being in focus.

Additionally, we provide the aligned sparse scene reconstruction from COLMAP with extracted camera poses, which can be used in a dense reconstruction pipeline. We also make available additional scenes with manually collected images and with plants that were captured only once.

\begin{table}[h]
    
    \centering \renewcommand{\arraystretch}{1.3}
    \begin{tabular}{m{0.12\linewidth}|m{0.07\linewidth} m{0.07\linewidth} m{0.12\linewidth} m{0.12\linewidth} m{0.07\linewidth}|m{0.06\linewidth}}
        Plant type & Bean & Corn & Soy bean (green) & Soy bean (yellow) & Sugar beet & Total\\
        \hline
        Plants & 2 & 5 & 3 & 2 & 7 & 19\\
        Scenes & 3 & 16 & 17 & 14 & 36 & 86\\
        Images & 280 & 1574 & 1699 & 1421 & 3570 & 8544\\
    \end{tabular}
    \caption{Dataset structure. The scenes where autonomously captured in the growing season 2024. Note that we additionally publish scenes which were manually captured in the growing season 2023.}
    \label{tab:dataset}
\end{table}

\begin{figure*}[htb!]
	\centering \newlength\teaserimghig\setlength\teaserimghig{2.75cm}
	\resizebox{1.0\linewidth}{!}{
	\includegraphics[]{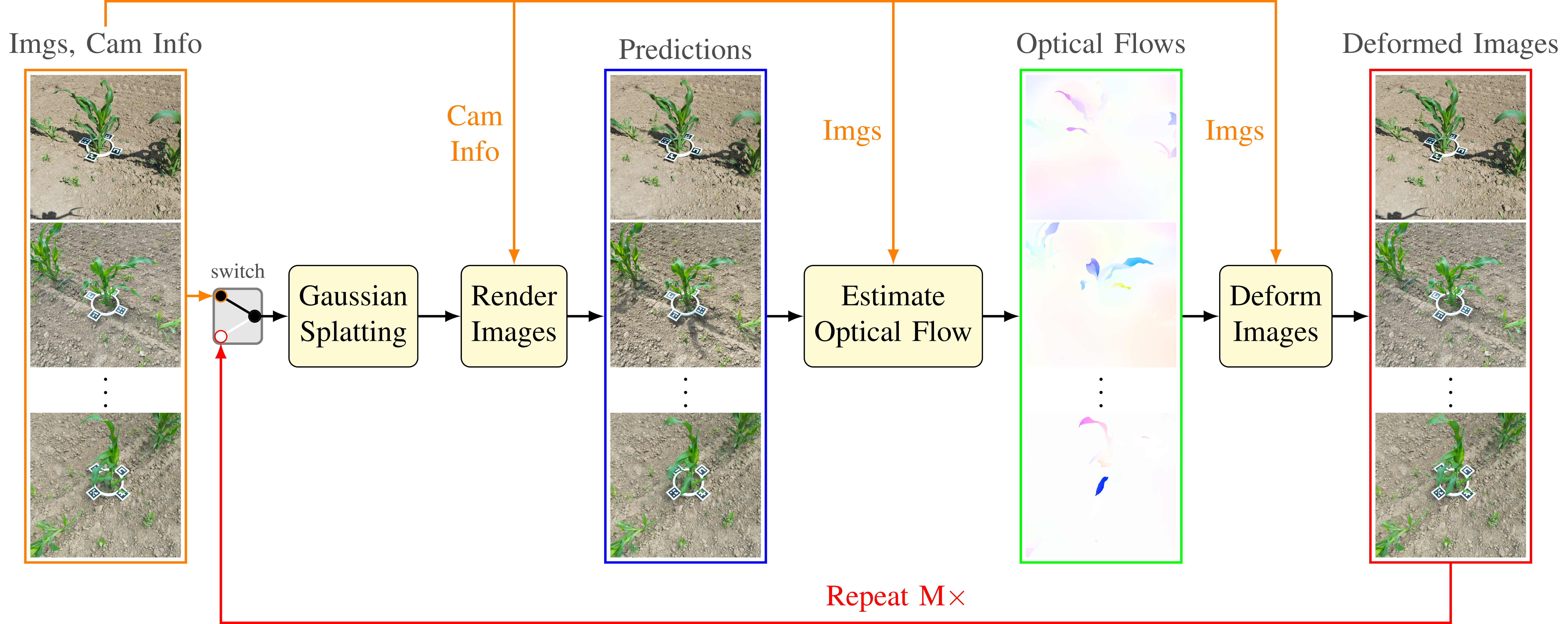} }

	\caption{Canonical 3D Reconstruction Pipeline. In the first iteration, the original input images are used to perform 3D Gaussian Splatting~\cite{kerbl20233d}. Subsequently, the input views are then rendered with the same camera parameters (Predictions). Using the predictions and original images, we then estimate the optical flow from the predictions to the original input images, which is then used to deform the input images into the current predictions (Deformed Images). In the following iterations, the 3D Gaussian Splatting is performed using the deformed images instead. These steps are repeated for a predefined number of iterations. Images cropped for visualization.} \label{fig:pipe:3d}
\end{figure*}

\section{3D Canonical Plant Reconstruction}
Our full 3D reconstruction pipeline, which is designed to align a non-rigid scene in a canonical representation, is illustrated in \cref{fig:pipe:3d} and explained in the following \cref{sub:3D:1,sub:3D:2,sub:3D:3}.
In agricultural fields, gusts of wind can cause plants to move. Furthermore, the UAV that sequentially captures images of the plant produces a significant amount of downwash as well.
This leads to non-corresponding leaf poses in the captured images. Our objective is therefore to estimate a canonical (motion free) 3D reconstruction from a non-rigid scene with unpredictable leaf motion.
Input to our method is a UAV-captured scene consisting of images and optimized camera parameters.

\subsection{Train 3D Gaussian Splatting (Step 1)} \label{sub:3D:1}
We train 3D Gaussian Splatting~\cite{kerbl20233d} with the original images captured by the UAV. 3D Gaussian Splatting represents the scene using a large number of Gaussians. These primitives are initialized with the sparse pointcloud generated by COLMAP. Each Gaussian is described with its position (mean), covariances (orientation and form), opacity, and color via spherical  harmonics. These parameters are optimized through differentiable rendering. Furthermore, 3D Gaussian Splatting~\cite{kerbl20233d} can add or remove Gaussians to achieve an optimal visual appearance.%

Given a set of $n$ images $I^t = \{I^t_1, I^t_2, ..., I^t_n\}$ at time step t, corresponding camera information $C = \{C_1, C_2, ..., C_n\}$ and a Gaussian Trainer $\Tilde{T}$, we therefore compute a Gaussian scene representation $\theta^t_{gs}$ with:
\begin{align}
\theta^t_{gs} = \Tilde{T}(I^t, C).
\end{align}
$I^{t=0}$ are initialized with ground truth images $I^{t=0} = I^{gt}$.

Note that our pipeline can also be used with neural rendering methods.

\subsection{Render Gaussian Scene from Input Views (Step 2)} \label{sub:3D:2}
The 3D Gaussians get projected by a tile-based rasterizer in order to compute 2D output images. A visability-aware $\alpha$-blending sorts all primitives to ensure correct depth layering~\cite{kerbl20233d}.
For a Gaussian Splatting Renderer $\Tilde{R}$, we render a set of predicted images $\hat{I}^t$ with the same camera information $C$ we used for training such that:
\begin{align}
    \hat{I}^t = \Tilde{R}(\theta^t_{gs}, C).
\end{align}
This yields a predicted image $\hat{I}^t_k$ for each ground truth image $I^{gt}_k$, such that $\hat{I}^t_k$ closely resembles $I^{gt}_k$.

\subsection{Estimate Optical Flow and Deform Images (Step 3)} \label{sub:3D:3}
We estimate optical flow from rendered images to corresponding ground truth images using RAFT~\cite{teed2020raft}. RAFT takes a pair of images $I_1$ and $I_2$ and computes a dense displacement field that maps each pixel in $I_1$ to a corresponding location in $I_2$. For every ground truth image $I^{gt}_k$ and its corresponding prediction image $\hat{I}^t_k$, we compute a dense displacement field
\begin{align}
    f_k = \mathcal{F}(\hat{I}^t_k, I^{gt}_k)\text{,}
\end{align}
where $\mathcal{F}$ downsamples the images, estimates the optical flow from $I^{gt}_k$ to $\hat{I}^t_k$, and finally upsamples the resulting flow to match the input image size.
Now we apply the displacement field $f_k$ to ground truth image $I^{gt}_k$ to deform it into the predicted image $\hat{I}^t_k$. For each pixel $(u, v)$ of $I^{gt}_k$, we get a new current deformed image $I^{t+1}_k$, with 
\begin{align}
    I^{t+1}_k(u,v) = I^{gt}_k((u,v) + f_k(u,v)). %
\end{align}

\subsection{Iterative Optimization} \label{sub:3D:it}
The steps described in \cref{sub:3D:1,sub:3D:2,sub:3D:3} are repeated for $M$ iterations using the updated image set $I^{t+1}$ as input for next iteration. 
Note that camera parameters $C$ are unchanged.

This iterative refinement regime compensates motion in the scene and gradually aligns the input images in a motion-free canonical representation. This results in sharper predictions and significantly better 3D reconstructions.

\subsection{Mesh Extraction} \label{sec:mesh}

We crop a region of interest for mesh extraction of $\theta_{gs}$. This is done by removing every primitive that is not inside a radius of $r$ to the center, where $r$ is chosen manually according to the size of the plant. With cropped scene $\Tilde{\theta}_{gs}$, we now render a set of cropped images $\Tilde{I}$ and optimize a 2D Gaussian volume~\cite{huang20242d} with them:
\begin{align}
\Tilde{I} = \Tilde{R}(\Tilde{\theta}_{gs}, C)
\end{align}
\begin{align}
\theta_{2Dgs} = \Tilde{T}(\Tilde{I}, C)
\end{align}

In contrast to 3D Gaussian Splatting methods, 2D Gaussian Splatting~\citep{huang20242d} uses flat 2D Gaussian disks which are placed directly on the object surface.
Because 2D Gaussian Splatting is explicitly modeling the surface of objects, it is better suited to extract a mesh.

The mesh extraction is done as proposed by~\citet{huang20242d} with a Marching Cubes voxelgrid resolution of $1536^3$. However, we optimize the texture in an separate step. 
We merge close vertices, decimate vertices to 500-700k~\citep{garland1997surface} and apply HC Laplacian Smoothing~\cite{vollmer1999improved} using Meshlab~\citep{cignoni2008meshlab}. We then extract a UV texture map in 8k resolution from the mesh with Blender~\citep{blender} Smart UV project using a angle limit of 25 degrees and a margin island of 0. 

We then optimize the texture UV map using a differentiable rendering approach. We draw a cropped image $\Tilde{I}_k$ and its corresponding camera information $C_k$. We then render an image $\Bar{I_k}$ from mesh $\hat{M}$ with camera parameter $C_k$. Finally, we compute the $L_1$ loss between $\Bar{I_k}$ and $\Tilde{I}_k$, backpropagate the loss to the texture map and optimize it. This yields a high-resolution texture map.

\section{Experiments} \label{sec:exp}

To evaluate our reconstruction pipeline, we conduct an ablation study and compare our method against other state-of-the-art approaches. We demonstrate that iterative optical flow compensation improves the quality of 3D Gaussian Splatting methods for our objective.

Due to scene motion, 3D Gaussian Splatting~\cite{kerbl20233d} tends to place Gaussian primitives near the camera. To avoid this kind of overfitting, we remove all Gaussian primitives with a z-value less than 30\,cm within a radius of 60\,cm around the camera during optimization.

All reported experiments are conducted with a image resolution of 2016x1512.

In the following, we refer to 3D Gaussian Splatting~\cite{kerbl20233d} as (GS) and to Deformable 3D Gaussian Splatting~\cite{yang2024deformable} as (DGS). Furthermore, GS+Ours-X or DGS+Ours-X refer to the respective baseline method combined with X iterations of our proposed optical flow compensation. Versions of \cref{fig:res1,fig:res2,fig:res4} with higher image quality can be found in the supplementary material.

\subsection{Qualitative Results}

\begin{figure*}[h]
	\centering
	\includegraphics[width=0.98\linewidth]{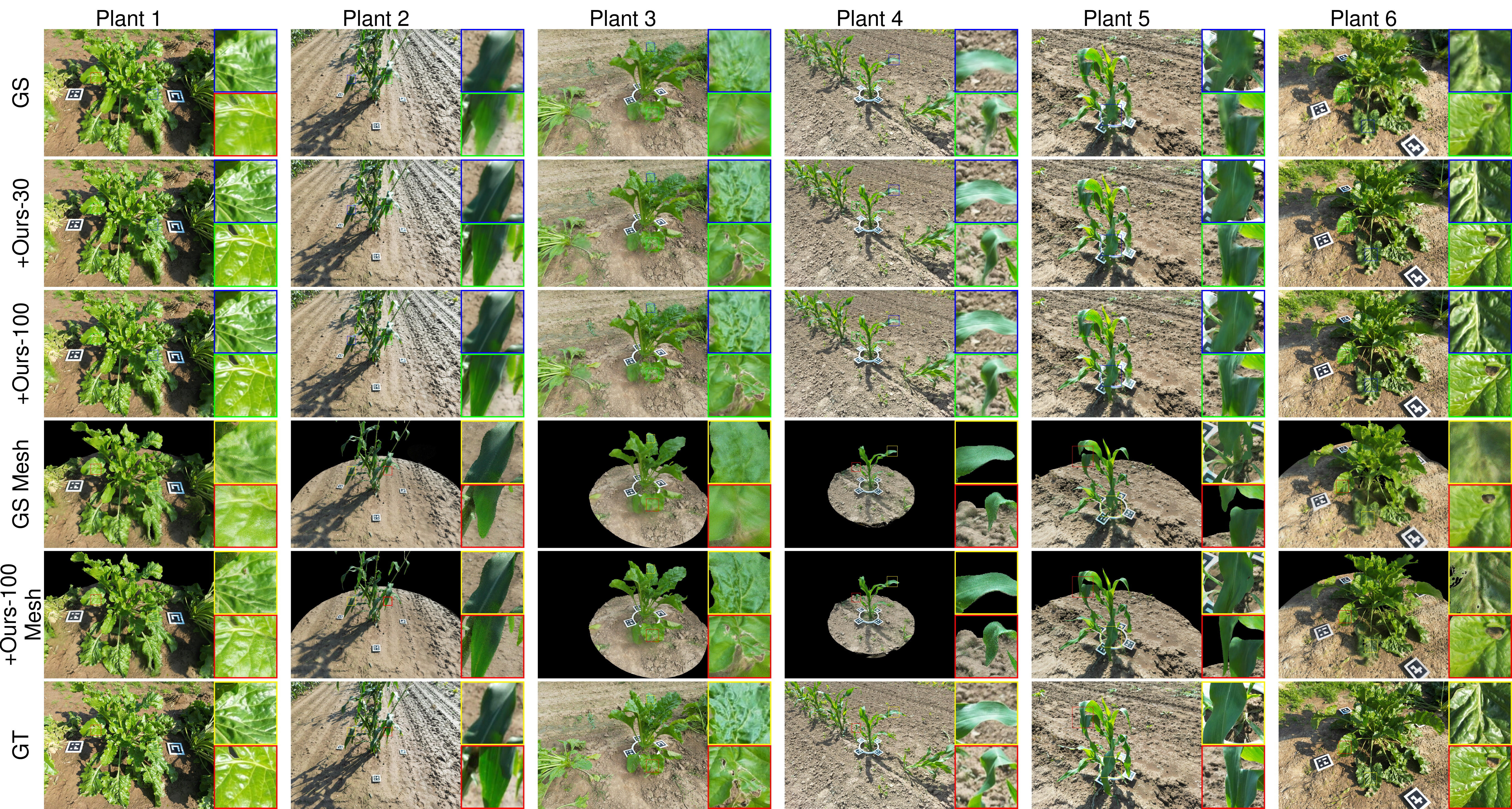}
	\caption{Comparison of 3D Gaussian Splatting~\cite{kerbl20233d} (GS) and our proposed method with optical flow compensation. The first row (GS) shows results from standard 3D Gaussian Splatting without motion compensation. The second and third rows (+Ours-30 and +Ours-100) show our results after 30 and 100 iterations of optical flow compensation, respectively. The fourth and fifth rows (GS Mesh and +Ours-100 Mesh) show meshes extracted from the 2D Gaussian representation before and after compensation. The last row (GT) presents the ground truth images for reference. Areas with notable improvements in visual quality are highlighted in the zoom-in boxes. Our approach leads to visibly sharper and more consistent textures across the scene, especially in regions affected by wind.}
	\label{fig:res1}
\end{figure*}

\subsubsection{3D Gaussian Splatting} \label{sub:3D:gaussian}
In order to investigate the benefits of our pipeline using methods that are not explicitly modeling motion in scenes, we compare against standard 3D Gaussian Splatting proposed by~\citet{kerbl20233d}.

We compare with 3D Gaussian Splatting, proposed by~\citep{kerbl20233d}, as the baseline method. In our experiments, we applied 100 iterations of our optical flow compensation step, requiring approximately four days on an NVIDIA A6000 GPU.

In \cref{fig:res1}, the effectiveness of our motion-compensating optical flow procedure becomes clearly visible. Across all tested scenes, we observe a consistent improvement in visual quality. The most significant improvement occurs between the baseline 3D Gaussian Splatting results and our method with 30 iterations. The improvements from 30 to 100 iterations are smaller compared to the first iterations. However, there are still refinements (see green spy box of Plant 4).

\subsubsection{Deformable 3D Gaussian Splatting}
Our approach is also compatible with other 3D reconstruction methods, such as Deformable 3D Gaussian Splatting proposed by~\citet{yang2024deformable}. They address non-static scenes by explicitly modeling motion by a deformation network. During the optimization process, the deformation network learns to adjust the Gaussian primitives in response to the observed motion, allowing for a more accurate representation of dynamic scenes.

To integrate DGS into our pipeline, we introduce a few modifications. First, we apply the removal of Gaussian primitives located too close to the camera, as described in \cref{sec:exp}. In addition, we define the first image as the canonical frame and apply our optical flow compensation only to the remaining images in the dataset, warping them into the corresponding reconstructions obtained with the time embedding of the canonical frame.

For the optical flow compensation within Deformable 3D Gaussian Splatting~\cite{yang2024deformable}, we limit the process to ten iterations, because the deformation network already captures a significant portion of the motion. Further iterations are not necessary and would significantly increase computation time. Note that ten iterations already require approximately four days on a NVIDIA A6000 GPU.

\begin{figure*}[h]
	\centering
	\includegraphics[width=0.98\linewidth]{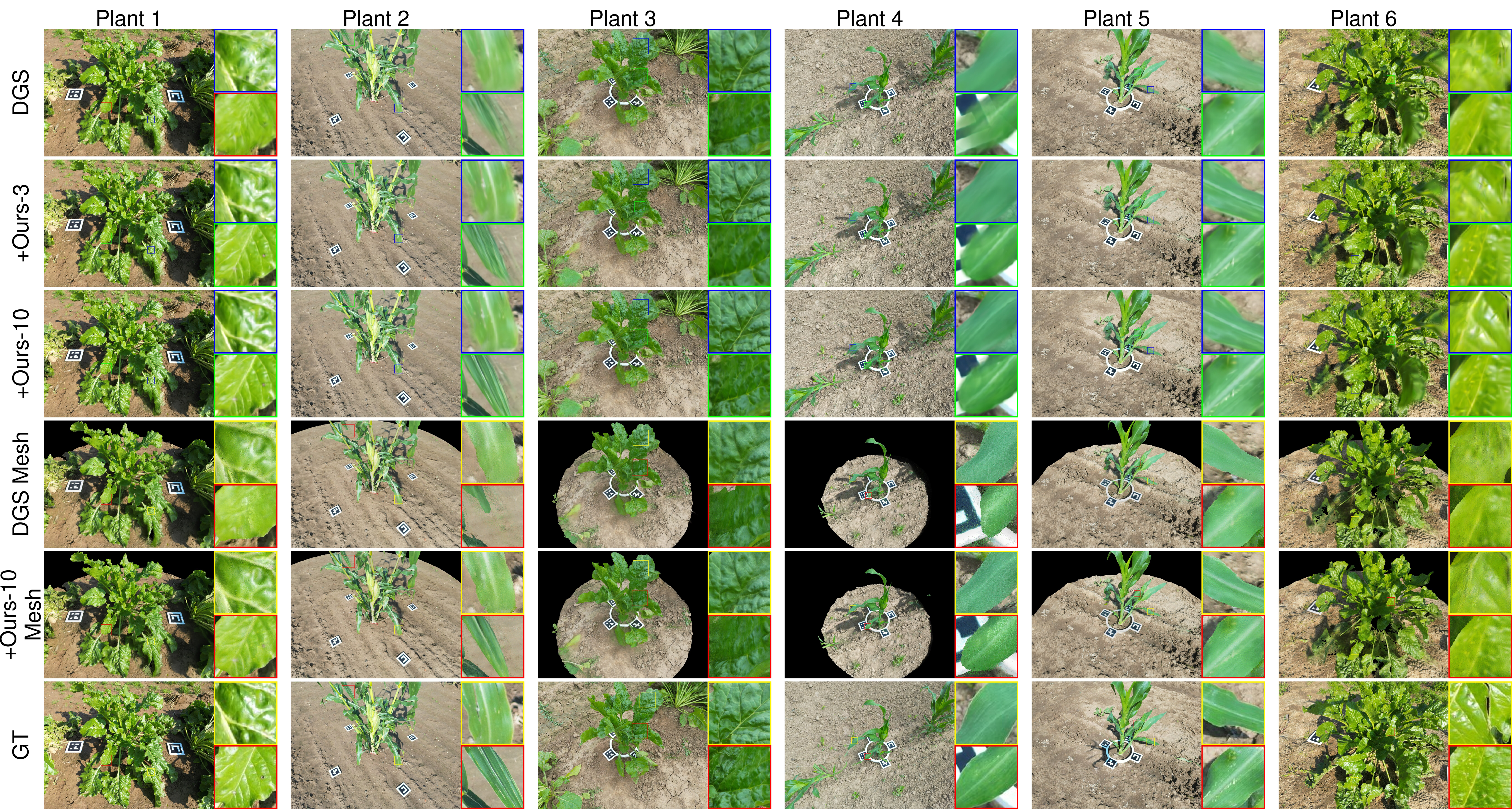} 
	\caption{Comparison of Deformable 3D Gaussian Splatting~\cite{yang2024deformable} (DGS) and our proposed method with optical flow compensation. As in figure \cref{fig:res1} first row (DGS) shows results from standard Deformable 3D Gaussian Splatting, second and third rows (+Ours-3 and +Ours-10) show our results after 3 and 10 iterations of optical flow compensation, fourth and fifth rows (DGS Mesh and +Ours-10 Mesh) show extracted meshes and the final row (GT) shows the ground truth. Our approach leads to some small improvements but not as significant as in the standard 3D Gaussian Splatting case.}
	\label{fig:res2}
\end{figure*}

As the results of \cref{fig:res2} show, our optical flow compensation improves the visual quality of baseline Deformable 3D Gaussian Splatting~\cite{yang2024deformable} in the reconstructed scenes. %
As expected, these improvements are smaller than in the case of standard 3D Gaussian Splatting~\cite{kerbl20233d}, which is also reflected in our quantitative evaluation in \cref{tab:quant}.

\subsection{Quantitative Results}

We present a quantitative evaluation in \cref{tab:quant}, using the mean PSNR, LPIPS, SSIM, and FID~\cite{heusel2017gans} scores computed across all input images from 11 randomly selected scenes.

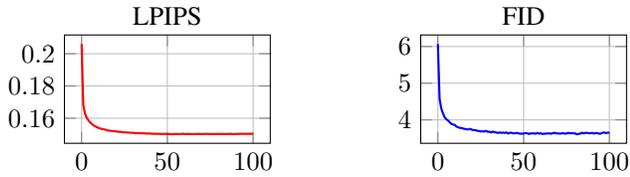
\begin{figure}[h]
\begin{tikzpicture}
\begin{groupplot}[
group style={
	group size=2 by 1,       %
	horizontal sep=2cm       %
},
width=0.5\linewidth,
height=0.35\linewidth,
grid=major
]

\nextgroupplot[
title={LPIPS},
title style={yshift=-2mm}
]
\addplot[thick, red, no markers] coordinates {
(0,0.20603940061547543) (1,0.16792515092952687) (2,0.16287929352711547) (3,0.16025161431594329) (4,0.15860636151649737) (5,0.15739308104596356) (6,0.15642950038340958) (7,0.15555822348052806) (8,0.15495393717153505) (9,0.1544231503185901) (10,0.15389481119811538) (11,0.15356685162267902) (12,0.153241599811749) (13,0.15307630549100312) (14,0.15282842843370004) (15,0.15252199756150897) (16,0.15226464691487226) (17,0.15220342406495055) (18,0.1520092445544221) (19,0.1519827597249638) (20,0.15186658500947736) (21,0.15163288465277713) (22,0.15158623564649715) (23,0.15148359440944414) (24,0.1513031302527948) (25,0.15121295926245776) (26,0.15119370543143965) (27,0.15112676778977566) (28,0.15102619906718082) (29,0.15100112240422855) (30,0.15098563375798138) (31,0.15079151381145822) (32,0.1507510635730895) (33,0.15077472694895483) (34,0.1506534690477631) (35,0.15062844849445603) (36,0.15070056697861714) (37,0.15054907938296144) (38,0.15050438479266384) (39,0.15052992705594412) (40,0.15049188883467154) (41,0.15039865955033085) (42,0.15037139248441564) (43,0.15029448278925636) (44,0.15030284061350604) (45,0.15032459865239534) (46,0.1502428384870291) (47,0.15023963599719783) (48,0.1502578700943427) (49,0.15023973469029772) (50,0.15018581439148296) (51,0.15012019131671298) (52,0.15014384979551487) (53,0.15004955046556215) (54,0.15011721535839817) (55,0.1501747943664139) (56,0.15018492420965976) (57,0.15012900425629183) (58,0.15016430707817727) (59,0.15034570409493012) (60,0.15016809704628858) (61,0.15013325658034196) (62,0.15022928090935406) (63,0.1501553284173662) (64,0.1501511448554017) (65,0.15011869771236722) (66,0.15018041861328213) (67,0.150151366550814) (68,0.15030374241146174) (69,0.15019401040944186) (70,0.15014555813913996) (71,0.15022527034309777) (72,0.1502591899376024) (73,0.15028014988384464) (74,0.1501994899457151) (75,0.15013005990873685) (76,0.150232458405874) (77,0.15021881990812042) (78,0.15017871400172061) (79,0.1502615967189724) (80,0.15020270153202794) (81,0.15023378484628416) (82,0.150243749584664) (83,0.15021941297433591) (84,0.15015319285744966) (85,0.15019971370019694) (86,0.1502379923178391) (87,0.15030083190311083) (88,0.15011925849047575) (89,0.1501606447385116) (90,0.150262614603747) (91,0.1502407409318469) (92,0.15027593529359862) (93,0.15032257574525748) (94,0.15031426736576992) (95,0.15033202984116292) (96,0.15029731300744142) (97,0.15036552998152644) (98,0.15035171411254192) (99,0.15030621983110903) (100,0.15045661762356757) 
};
\node at (rel axis cs:0.5,1) [anchor=south] {SSIM};

\nextgroupplot[
title={FID},
title style={yshift=-2mm}
]
\addplot[thick, blue, no markers] coordinates {
(0,6.068920774952556) (1,4.568364744290843) (2,4.309693740638225) (3,4.174094205826446) (4,4.067333501119613) (5,4.017243703006797) (6,3.979378782719992) (7,3.931592446915427) (8,3.888494947445336) (9,3.873569783835947) (10,3.858270106785648) (11,3.8133686956226343) (12,3.796474296199949) (13,3.780291011350849) (14,3.7778877565602516) (15,3.7611812216511287) (16,3.7487698421283624) (17,3.744557574612168) (18,3.7413535707955634) (19,3.754875691740703) (20,3.7286472762959146) (21,3.7213107100762737) (22,3.7188435120462042) (23,3.7024003644313095) (24,3.697133701377425) (25,3.684965346033179) (26,3.6895166911986546) (27,3.6819422830895974) (28,3.7015450574833064) (29,3.675454922299353) (30,3.6777007825343215) (31,3.6572986226262554) (32,3.68185324598787) (33,3.6656125028736253) (34,3.657688700554205) (35,3.643970162578407) (36,3.6561966656483618) (37,3.6465248163558197) (38,3.653528947124634) (39,3.6448904965272075) (40,3.6576662735094403) (41,3.6282679928958186) (42,3.6482651479235217) (43,3.647846649649648) (44,3.6322068610338745) (45,3.637311822034533) (46,3.645262542848765) (47,3.6327833779981145) (48,3.622654984617064) (49,3.6231696719036246) (50,3.6298546809894305) (51,3.6373674487349263) (52,3.637050029110847) (53,3.6169703286993724) (54,3.6306448382195864) (55,3.6302103443867537) (56,3.6343801957701674) (57,3.6218681249498106) (58,3.644223438695381) (59,3.6268220726665117) (60,3.633351977094792) (61,3.6183273794180226) (62,3.620673973982625) (63,3.6295374170528656) (64,3.6290539408649494) (65,3.630232510103768) (66,3.6235694133772607) (67,3.636717320066411) (68,3.6485182562539453) (69,3.6334555324320466) (70,3.6291360864508433) (71,3.6271023999950924) (72,3.6298066325845686) (73,3.6381944144160854) (74,3.6239676494102038) (75,3.629927718036235) (76,3.6452434040730837) (77,3.6341677479046854) (78,3.641749281623679) (79,3.632416471356353) (80,3.632715661995342) (81,3.60848200598313) (82,3.617121789412584) (83,3.6267819978042484) (84,3.650182696332325) (85,3.631037395341217) (86,3.647663043616502) (87,3.6458133518558267) (88,3.633983888366758) (89,3.6383711914794246) (90,3.6452893707736846) (91,3.6279004017543515) (92,3.6426932048543037) (93,3.639274185284038) (94,3.6239065874283725) (95,3.625908070178717) (96,3.6498616968852517) (97,3.6557181434593655) (98,3.6409762158187955) (99,3.650864310357234) (100,3.654742301725723) 
};

\end{groupplot}
\end{tikzpicture} 
\caption{LPIPS and FID scores (y-axis) of GS+Ours plotted after each iteration (x-axis), following the quantitative evaluation from \cref{tab:quant}.} \label{fig:plot} 
\vspace{-1ex}
\end{figure}

The FID score is calculated between the ground truth input images and the reconstructed images. Although these pairs often do not align perfectly due to leaf motion in the scene, a lower FID indicates that the distribution of generated images more closely matches the distribution of real images in the feature space.
In contrast, PSNR, SSIM, and LPIPS assume a direct correspondence between the generated and ground truth images. Since our method reconstructs a canonical scene that compensates for leaf motion, direct comparison with the ground truth is not feasible. To address this, we use optical flow to deform the generated images into the viewpoint of the ground truth, and evaluate the deformed predictions. The accuracy of the optical flow improves when the predicted image closely resembles a spatially deformed version of the ground truth image at the same camera pose.
As shown in \cref{tab:quant}, our method significantly outperforms the baseline methods across all evaluation metrics. \cref{fig:plot} shows that our method yields the greatest improvements in the initial iterations. This demonstrates that our method already provides significant benefits with fewer iterations and shorter training times.  

\begin{table}[b]
	
	\centering 
	\begin{tabular}{m{0.18\linewidth}|m{0.08\linewidth} m{0.08\linewidth}m{0.08\linewidth} m{0.08\linewidth}}
		Method   & PSNR$\uparrow$ & LPIPS$\downarrow$ & SSIM$\uparrow$ & FID$\downarrow$\\
		\toprule
		
		DGS~\citep{yang2024deformable}      & 24.15& 0.1670& 0.8160 & 3.8830 \\
		{\,\,\,}+Ours-10  & 24.31& 0.1530& 0.8395 & \textbf{3.5613} \\ \midrule
		GS~\citep{kerbl20233d}      & 24.41& 0.2060& 0.7895 & 6.0689\\
		{\,\,\,}+Ours-100 & \textbf{25.37}& \textbf{0.1505} & \textbf{0.8502} & 3.6547\\
		
	\end{tabular}
	\caption{Quantitative evaluation of our method compared with the two baseline methods DGS~\citep{yang2024deformable} and GS~\citep{kerbl20233d}.} \label{tab:quant}
\end{table}

\subsection{Limitations in Mesh Extraction} \label{sub:mesh}
\begin{figure*}[h]
	\centering
	\includegraphics[width=0.92\linewidth]{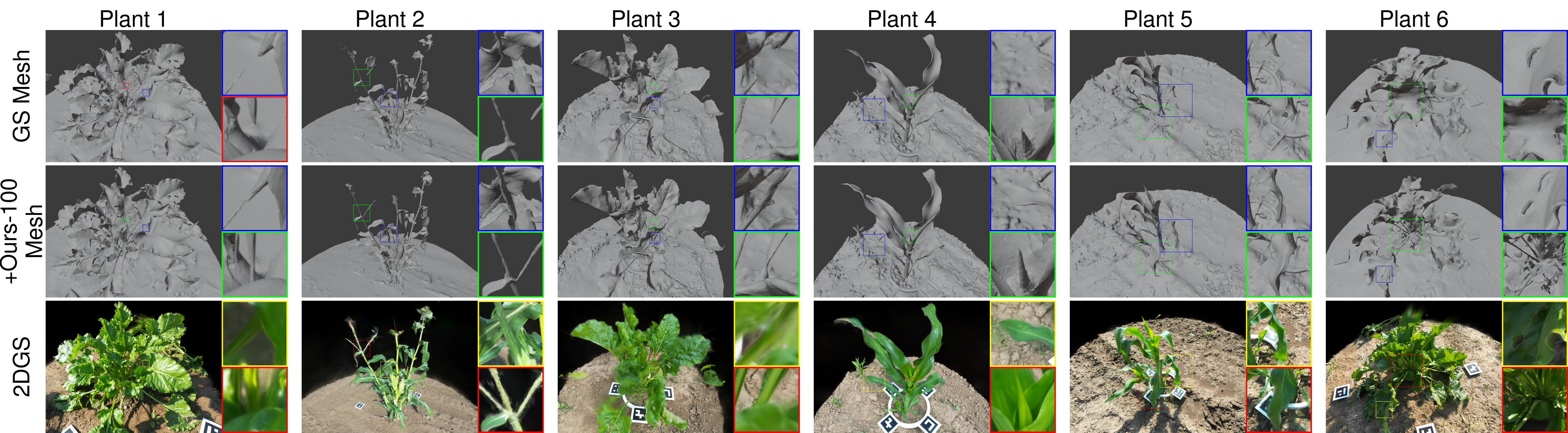} \vspace{-0.5ex}
	\caption{We compare geometry of GS~\cite{kerbl20233d} (first row) and GS combined with our proposed method after 100 iteration (second row). Meshes are extracted using 2D Gaussian Splatting~\cite{huang20242d} (2DGS) as described in \cref{sec:mesh}. The third row shows the rendered image of the 2DGS volume before extracting the mesh of GS+Ours-100 (second row). The zoom-in regions highlight improvements in geometry. As discussed in \cref{sub:mesh}, certain fine structures such as thin plant stems recovered by our method remain absent in the reconstructed meshes. This limitation is inherent to mesh extraction using 2DGS.}
	 \vspace{-2ex}
	\label{fig:res4}
\end{figure*}

While our method significantly enhances the texture quality of reconstructed scenes (see \cref{fig:res1,fig:res2}), its impact on geometry is more modest, but still evident (see \cref{fig:res4}). Notably, the geometry of stems and leaves that appear blurry or are missing in the baseline method is better recovered when meshes are extracted after applying our proposed optical flow compensation. Since geometry extraction relies on 2D Gaussian Splatting~\cite{huang20242d}, which is inherently imprecise, not all improvements introduced by our method are faithfully reflected in the extracted meshes. For example, a very thin stem accurately recovered by our approach may still be poorly captured during mesh extraction. We therefore argue that more accurate mesh extraction techniques are required to fully exploit the benefits of our method.

\section{Conclusion}
We presented a 3D reconstruction method that builds on SOTA techniques to create a high-resolution reconstruction from plant images containing a significant amount of motion. Especially the downwash from the UAV creates significant plant motion in the sequentially captured images. Our method improves the results of both rigid SOTA 3D reconstruction methods and SOTA methods that explicitly model motion. In addition, we release a dataset of agricultural plants collected with a self-developed Android application controlling on a small and affordable UAV. Future work includes improving mesh extraction techniques, refining optical flow estimation, and reducing training time.

\section{Acknowledgment}
This work was funded by the Deutsche Forschungsgemeinschaft (DFG, German Research Foundation) under Germany’s
Excellence Strategy–EXC 2070–390732324.
\printbibliography

\end{document}